%% file: lrec2026-example.tex
\documentclass[10pt, a4paper]{article}

\usepackage{booktabs}
\usepackage[final]{lrec2026} 

\input{package_stuff}

\title{\textsc{Talk2Ref}: A Dataset for Reference Prediction from Scientific Talks}

\name{Frederik Broy, Maike Züfle, Jan Niehues} 

\address{Karlsruhe Institute of Technology, Germany \\
         frederik.broy@student.kit.edu,\\
         \{maike.zuefle, jan.niehues\}@kit.edu\\}

\abstract{
Scientific talks are a growing medium for disseminating research, and automatically identifying relevant literature that grounds or enriches a talk would be highly valuable for researchers and students alike.
We introduce Reference Prediction from Talks (RPT), a new task that maps long, and unstructured scientific presentations to relevant papers. To support research on RPT, we present \textsc{Talk2Ref}, the first large-scale dataset of its kind, containing 6,279 talks and 43,429 cited papers (26 per talk on average), where relevance is approximated by the papers cited in the talk’s corresponding source publication. 
We establish strong baselines by evaluating state-of-the-art text embedding models in zero-shot retrieval scenarios, and propose a dual-encoder architecture trained on \textsc{Talk2Ref}. We further explore strategies for handling long transcripts, as well as training for domain adaptation. Our results show that fine-tuning on \textsc{Talk2Ref} significantly improves citation prediction performance, demonstrating both the challenges of the task and the effectiveness of our dataset for learning semantic representations from spoken scientific content. The dataset and trained models are released under an open license to foster future research on integrating spoken scientific communication into citation recommendation systems.
 \\ \newline \Keywords{Scientific Talks, Citation Prediction, Spoken Language Processing} }

\begin{document}

\maketitleabstract

\input{content}

\section{Acknowledgements}
Part of this work received support from the European Union’s Horizon research and innovation programme under grant agreement No 101135798, project Meetween (My Personal AI Mediator for Virtual MEETtings BetWEEN People).

\section{Bibliographical References}\label{sec:reference}

\bibliographystyle{lrec2026-natbib}
\bibliography{lrec2026-example}

\section{Language Resource References}\label{lr:ref}
\bibliographystylelanguageresource{lrec2026-natbib}
\bibliographylanguageresource{languageresource}

\clearpage
\input{appendix}

\end{document}

%% file: package_stuff.tex
\usepackage{amsmath}
\usepackage{adjustbox}

\usepackage[capitalize]{cleveref}
\usepackage{graphicx}
\usepackage{xspace}  
\usepackage{twemojis}
\usepackage{xcolor}
\usepackage{xurl}
\usepackage{footmisc}
\usepackage{array} 

\usepackage[acronym]{glossaries}
\newacronym{dpr}{DPR}{Dense Passage Retrieval}
\newacronym{acl}{ACL}{Association for Computational Linguistics}
\newacronym{api}{API}{Application Programming interface}
\newacronym{ai}{AI}{Artificial Intelligence}
\newacronym{tf-idf}{TF-IDF}{Term Frequency-Inverse Document Fequency}
\newacronym{svm}{SVM}{Support Vector Machine}
\newacronym{resca}{Resca}{Recommender System for Scientific Articles}
\newacronym{url}{URL}{Uniform Resource Locator}
\newacronym{asr}{ASR}{Automatic Speech Recognition}
\newacronym{grobid}{GROBID}{GeneRation Of Bibliographic Data}
\newacronym{sag}{SAG}{Speech-to-Abstract Generation}
\newacronym{nlp}{NLP}{Natural Language Processing}
\newacronym{snli}{SNLI}{Stanford Natural Language Inference}
\newacronym{multinli}{MultiNLI}{Multi-Genre Natural Language Inference}
\newacronym[plural=PDFs, firstplural={Portable Document Formats (PDFs)}]{pdf}{PDF}{Portable Document Format}
\newacronym[plural=RNNs, firstplural={Recurrent Neural Network (RNNs)}]{rnn}{RNN}{Recurrent Neural Network}
\newacronym{faiss}{Faiss}{Facebook AI Similarity Search}
\newacronym{bert}{BERT}{Bidirectional Encoder Representations from Transformers}
\newacronym{aan}{AAN}{ACL Anthology Network}
\newacronym{aclarc}{ACL-ARC}{ACL Anthology Reference Corpus}
\newacronym[plural=LLMs, firstplural={Large Language Models (LLMs)}]{llm}{LLM}{Large Language Model}

\newacronym{tfidf}{TF-IDF}{Term Frequency–Inverse Document Frequency}
\newacronym{cata++}{CATA++}{Collaborative Dual Attentive Autoencoder}
\newacronym{scibert}{SciBERT}{Scientific Bidirectional Encoder Representations from Transformers}
\newacronym{dpcnn}{DPCNN}{Deep Pyramid Convolutional Neural Network}
\newacronym{qa}{QA}{Question Answering}
\newacronym{doi}{DOI}{Digital Object Identifier}

\DeclareRobustCommand{\microtext}{%
  \raisebox{-0.3\height}{\includegraphics[height=1.3em]{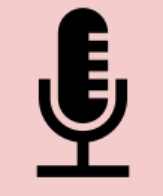}}%
  \xspace
}

\DeclareRobustCommand{\doctext}{%
  \raisebox{-0.3\height}{\includegraphics[height=1.3em]{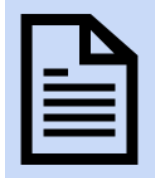}}%
  \xspace
}

\newcommand{\microphone}{%
  \raisebox{-0.3\height}{\includegraphics[height=1.7em]{microphone.pdf}}%
  \xspace
}

\newcommand{\docicon}{%
  \raisebox{-0.3\height}{\includegraphics[height=1.7em]{doc.pdf}}%
  \xspace
}
\newcommand{\robot}{\scalebox{1.25}{\twemoji{robot}}}
\newcommand{\down}{\scalebox{1.25}{\twemoji{fast down button}}}

\newcommand{\vpad}{\rule{0pt}{18pt}}

%% file: content.tex
\input{01_introduction}

\input{2_related_work.tex}
\input{3_dataset.tex}

\section{Analysis} \label{sec:analysis}
We now use our dataset, \textsc{Talk2Ref}, to benchmark baseline models for reference prediction from scientific talks. This analysis serves two purposes: first, to assess the difficulty of the task, and second, to evaluate whether models can be effectively trained on the dataset, thereby establishing its value as a resource for developing reference prediction systems.

\input{4_experiments}

\input{5_results.tex}

\section{Conclusion} \label{sec:conclusion}
This work introduces the task of Reference Prediction from Talks (RPT). RPT is practically relevant, as researchers, lecturers, and students could benefit from automated recommendations of related works for scientific talks. At the same time, it is highly challenging, requiring models to map long, noisy, and unstructured spoken content to the corresponding relevant papers.

To support this task, we present \textsc{Talk2Ref}, the first large-scale benchmark for RPT. To our knowledge, it is also the first dataset to incorporate the spoken modality on the query side of a citation prediction task.

Despite the challenges, our experiments show that RPT is tractable. By training dual-encoder architectures inspired by dense passage retrieval, we demonstrate that models can effectively learn semantic representations aligning scientific talks with their cited references. Finetuning these architectures on \textsc{Talk2Ref} significantly improves performance over zero-shot and heuristic baselines, confirming that dedicated training is essential for this modality.

We will release the dataset and the trained models under an open license, providing the community with ready-to-use tools for this task. We hope that \textsc{Talk2Ref} and the accompanying models will serve as a foundation for further research at the intersection of speech, language, and scholarly retrieval. 

Future work may explore audio-based representations of talks, multimodal fusion of speech and text, and the development of encoders capable of handling longer input sequences, improving the ability to accurately predict relevant citations from spoken scientific content.

\section{Ethical Considerations}
The \textsc{Talk2Ref} dataset is constructed from publicly available scientific talks and papers, and does not include any private or sensitive information. As such, we do not anticipate significant ethical risks associated with its release. All content is already in the public domain and intended for scholarly use.

Potential considerations include:
\begin{itemize}
    \item \textbf{Biases in scientific citations:} Like all citation datasets, \textsc{Talk2Ref} reflects the citation practices of the underlying field, which may underrepresent certain authors, institutions, or geographic regions. Models trained on this data may inadvertently perpetuate these biases.
    \item \textbf{Misuse for evaluation:} the dataset is intended for research on reference prediction and related tasks. Misuse for ranking or evaluating researchers or institutions should be avoided.
\end{itemize}

Overall, we consider \textsc{Talk2Ref} suitable for research use, while encouraging users to remain aware of the inherent limitations and potential biases of citation-based datasets.

%% file: 01_introduction.tex
\section{Introduction}\label{sec:introduction}
\footnotetext[1]{\label{fn:Talk2Ref}The dataset is available at \url{https://huggingface.co/datasets/s8frbroy/talk2ref}.}
In recent years, the number of recorded scientific talks across conferences, academic platforms, and educational settings has increased dramatically. These recordings represent a rapidly growing source of scientific communication, enabling researchers, students, and practitioners to revisit presentations, lectures, and discussions, and engage with scientific ideas beyond traditional written publications. 
However, identifying related or thematically relevant work from a scientific \textit{talk} remains challenging and time-consuming, yet access to such related content is highly valuable for researchers who wish to explore prior work, follow up on mentioned ideas, and discover new connections.

\begin{figure}[t]
    \centering
    \includegraphics[width=1\linewidth]{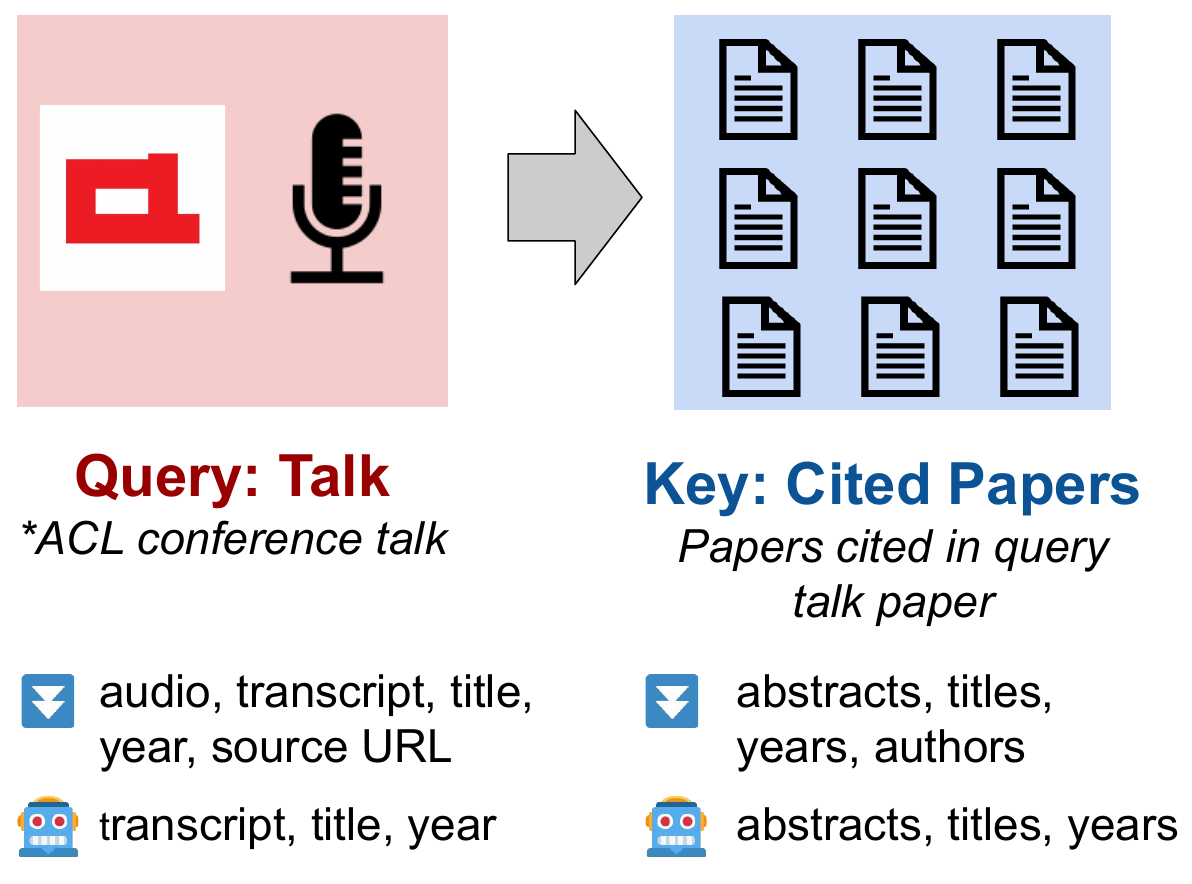}
    \caption{Illustration of the \textbf{\textsc{Talk2Ref}} dataset and its use in the task of Reference Prediction from Scientific Talks (RPT), where query talks are paired with their cited papers. The \down\phantom{1}represents the information included in \textsc{Talk2Ref}, and \robot\phantom{0}represents the input used by our model for predicting cited papers.}
    \label{fig:citation_prediction}
\end{figure}

Reference or citation prediction has been extensively studied in the context of written scientific text, in both \textit{local} settings, where models predict in-text references based on the surrounding context \citep{zhang-ma-2020-dual, gu2022localcitationrecommendationhierarchicalattention, çelik2025citebartlearninggeneratecitations, RS_dual_att}, and \textit{global} settings, which recommend relevant references for an entire document (\citealplanguageresource{RS_content_important}; \citealp{RS_resca, RS_clustering}). In both cases, methods assume clean, formal, and linguistically organized inputs that capture the semantics of scientific discourse, allowing models to rely on concise and well-structured representations, such as abstracts, titles, or surrounding sentences, without processing the full text.

By contrast, scientific talks are unstructured and often noisy, containing disfluencies, filler expressions, and spontaneous language \citeplanguageresource{nutshell} that lacks the precision of academic prose. Moreover, transcripts are typically long and verbose, spanning tens of minutes, which poses additional challenges for effective representation, retrieval, and modeling. As a result, the entire talk must be processed to capture its content, unlike written papers where state-of-the-art methods rely on abstracts and section headings (\citealplanguageresource{RS_content_important}; \citealp{RS_resca, RS_clustering}).

Importantly, this introduces not only a domain shift from written to spoken language, but also an inherent mismatch between the query and retrieval spaces in style and domain: the queries are derived from spontaneous spoken content, while the targets are formal written papers. Capturing semantic correspondences across these distinct modalities and linguistic styles is therefore substantially more difficult, rendering traditional text-based citation prediction models poorly suited for this setting.

To address this gap, we introduce the novel task of Reference Prediction from Talks (RPT): given a scientific talk as a query, the \textit{query talk}, the goal is to predict the set of papers that are relevant to the talk, the \textit{relevant papers}.  RPT extends the paradigm of citation recommendation from written to spoken scientific communication, opening new opportunities for integrating spoken research content into digital scholarly systems. An overview of 
the reference prediction setting is illustrated in \cref{fig:citation_prediction}.

To the best of our knowledge, no existing dataset supports research on this problem. We therefore construct and release \textsc{Talk2Ref}, the first large-scale dataset that pairs scientific presentations with their corresponding relevant papers, modeling relevance using the papers referenced in each talk’s source publication. \textsc{Talk2Ref} includes 6,279 talks and 43,429 papers, with an average of 26 references per talk, providing a foundation for systematically studying reference prediction from spoken scientific content at scale.

To provide reference points for future research, we establish competitive baseline models. We begin by evaluating state-of-the-art sentence embedding models, \citep{SentenceBERT, specter} in a zero-shot retrieval setting, where both talks and papers are encoded into a shared embedding space, and the most semantically similar papers are retrieved for each talk.

We then fine-tune a dual-encoder model \citep{DPR} on \textsc{Talk2Ref} to better align representations of talks and their cited papers. This adaptation enables the model to learn task-specific associations beyond those captured by pretrained embeddings. A key challenge in this setting is the considerable length and complexity of talk transcripts, which exceed the typical input limits of Transformer-based encoders \citep{bert, SentenceBERT, specter}. To handle this, we design strategies that enable the model to represent long talks effectively while preserving their overall semantics.

Our results show that fine-tuning on \textsc{Talk2Ref} yields substantial improvements over zero-shot performance, demonstrating that the dataset effectively supports model adaptation and generalization. Together, these findings indicate that \textsc{Talk2Ref} not only provides a new research task, but also offers strong, competitive benchmarks that future models can build upon.

The main contributions of this work are threefold:
\begin{enumerate}
    \item We introduce and publicly release the \textsc{Talk2Ref} dataset for the new task of RPT under an open license (CC-BY-4.0).\textsuperscript{\ref{fn:Talk2Ref}}
    \item We establish strong baselines and propose a dual-encoder framework for modeling citation prediction from scientific talks, which we also release under an open license.\footnote[2]{\url{https://huggingface.co/s8frbroy/talk2ref_query_talk_encoder} and \url{https://huggingface.co/s8frbroy/talk2ref_ref_key_cited_paper_encoder}.}
    \item We analyse different aggregation mechanisms and training scenarios for RPT.
\end{enumerate}
 

%% file: 2_related_work.tex
\section{Related Work}\label{sec:rel_work}

Citation recommendation is commonly divided into two tasks: \textit{local} and \textit{global} citation prediction. Local approaches predict in-text references based on surrounding context \citep{zhang-ma-2020-dual, gu2022localcitationrecommendationhierarchicalattention, çelik2025citebartlearninggeneratecitations, RS_dual_att}, while global approaches suggest relevant references for an entire document (\citealplanguageresource{RS_content_important}; \citealp{RS_resca, RS_clustering}). As our work targets reference prediction from entire talks, we focus exclusively on the global task.

\paragraph{Datasets.} Unlike local citation tasks—which benefit from standardized benchmarks such as FullTextPeerRead~\citeplanguageresource{FullTextPeerRead} or RefSeer~\citeplanguageresource{RefSeer}, global citation recommendation lacks widely used datasets. Most prior work instead builds on large-scale bibliographic corpora, especially the DBLP-Citation-Network and ACM-Citation-Network \citeplanguageresource{dblp_acm}, both provided by AMiner\footnote[3]{\url{https://www.aminer.cn/citation}}. These datasets contain millions of papers and citation links in the domain of computer science, enriched with metadata such as title, abstract, authorship, venue, publication year, and publisher.

Variants or subsets of these corpora are widely used to train and evaluate citation recommendation models. Prior studies rely on smaller subsets of the DBLP or ACM-Citation-Network containing tens of thousands of papers (\citealp{RS_resca}; \citealplanguageresource{RS_content_important}), as well as larger-scale versions comprising hundreds of thousands of documents and citation links~\citep{RS_clustering, personalized_cit, global_citation_rec_gen_ad}.

Other datasets also complement these large-scale bibliographic corpora. The ACL Anthology Network ~\citeplanguageresource[AAN]{AAN} provides a domain-specific benchmark for natural language processing research and is frequently reused for citation recommendation tasks~\citep{personalized_cit, global_cit_multi_view}. Broader, cross-domain resources such as the STM-Knowledge Graph~(\citealplanguageresource{STM_KG}; \citealp{cit_rec_papers_kg}, STM-KG), OpenCorpus~\citeplanguageresource{RS_content_important}, and combined datasets built from the Microsoft Academic Graph and CiteSeerX~(\citealplanguageresource{citeseerx, microsoft}; \citealp{global_cit_knowledge_gra}) have also been employed to explore large-scale citation and knowledge graph modeling.

Despite this variety, all existing datasets focus exclusively on written-scientific-text papers, abstracts, or titles, leaving spoken scientific content entirely unexplored. Our dataset, \textsc{Talk2Ref}, addresses this gap by pairing scientific talks with their cited papers, enabling research on reference prediction from talks.

\paragraph{Modeling Approaches.}  Architectures for global citation recommendation differ primarily in the type of input they consider. Some rely only on document titles~\citep{RS_resca}, others use titles and abstracts (\citealplanguageresource{RS_content_important}; \citealp{RS_clustering}), and some leverage full-text papers~\citep{RS_Svm}. Additional information such as citation graphs~\citep{RS_resca, RS_clustering} or user profiles~\citep{RS_clustering} has also been incorporated to improve prediction.
These systems can be broadly categorized into classification-based and retrieval-based architectures. Classification-based models~\citep{RS_Svm, RS_clustering} predict citation relevance for each query–candidate pair, while retrieval-based models (\citealp{RS_resca}; \citealplanguageresource{RS_content_important}) precompute embeddings for all candidates and rank them by similarity to the query, allowing for more efficient large-scale recommendation.

%% file: 3_dataset.tex
\section{Task Formulation and Dataset}\label{sec:dataset}
\input{tables/dataset_stats}
Global Reference Prediction from Scientific Talks (RPT) aims to identify the set of relevant references for a talk, bridging unstructured spoken content and structured scientific literature. Formally, given a scientific presentation, the \microtext \textit{query talk}, $T$, the objective is to retrieve a set of \doctext \textit{relevant papers} $\{R_1, R_2, \dots, R_n\}$,  where $n$ may differ across talks, reflecting the varying number of references associated with each presentation. To support this task, we introduce \textsc{Talk2Ref}, which, to the best of our knowledge, is the first dataset of its kind.

\subsection{Dataset Construction}
To support Reference Prediction from Talks (RPT), a dataset must pair each query talk, the audio recording of a scientific presentation, with its relevant papers. In \textsc{Talk2Ref}, we model relevance using the citations in the source publication associated with each talk. Each input sample includes the talk’s audio, transcript, title, publication year, and source URL, while each cited paper is represented by its title, abstract, and metadata such as authorship and year.

The Association for Computational Linguistics (ACL) Anthology\footnote{\url{https://aclanthology.org}} provides a suitable source, offering talks linked to papers with accessible citation information. All talks are distributed under the Creative Commons Attribution 4.0 International License, allowing us to use them to construct a new dataset. To streamline construction, we build on the NUTSHELL dataset~\citeplanguageresource{nutshell}, which aggregates ACL conference talk recordings with their corresponding paper abstracts and metadata. Using this foundation, we compile \textsc{Talk2Ref}, a dataset that meets the requirements of RPT by pairing query talks with the references of their underlying papers.

However, while NUTSHELL provides talk recordings and paper metadata, it does not include the list of references from the query talk papers. We construct \textsc{Talk2Ref} by extracting the references from the original papers to obtain the full set of cited papers for each talk. The dataset construction process is summarized as follows:

\begin{enumerate}
    \item  \textbf{\microtext PDF retrieval.} Each query talk in NUTSHELL provides a link to its ACL page; from there we obtain the PDF of the corresponding paper, discarding any corrupted files. 
    \item \textbf{\microtext\doctext Query talk paper parsing.} We parse the PDF of each query talk paper using GeneRation Of Bibliographic Data ~\citep[][GROBID]{GROBID}. From the parsed document, we extract the paper title (for the query talk) as well as the structured metadata of the references, including titles, author lists, venues, years of publication, and Digital Object Identifiers (DOIs). The talk’s audio and year are obtained from the original metadata.
    \item \textbf{\microtext Transcript generation.} We transcribe the audio of each query talk to produce textual input for processing, using \texttt{whisper-large-v3}~\citep{whisper}.
    \item \textbf{\doctext Abstract retrieval for cited papers.} Since cited papers are not exclusively from ACL and coverage varies across sources, we use the metadata obtained in Step 2 for each cited paper to query six bibliographic APIs and datasets to obtain abstracts, including  
    Crossref\footnote{\url{https://www.crossref.org/documentation/retrieve-metadata/rest-api/}}, arXiv\footnote{\url{https://info.arxiv.org/help/api/basics.html}}, OpenAlex\footnote{\url{https://openalex.org/rest-api}}, Semantic Scholar ~\citeplanguageresource{semantic_scholar}, the  Laion arXiv-abstract dataset\footnote{\url{https://huggingface.co/datasets/laion/arXiv-abstract}}, and the  ACL OCL corpus~\citeplanguageresource{acl_ocl}.
    We prioritize DOI-based queries, as DOIs provide a unique and unambiguous identifier for each paper; however, they are not always present in the parsed metadata. Therefore, we first perform title–author queries, which may return abstracts directly or allow us to recover a DOI. Once a DOI is obtained, it is used for subsequent API queries, reducing the risk of mismatches due to title variations. 
    \item \textbf{\doctext Post-processing.} We filter out incorrect or placeholder abstracts, such as entries containing only authorship, venue, or year information.
    
\end{enumerate}

The resulting dataset links each query talk, represented by its audio, transcript, title, year, and source URL to its cited papers, which are enriched with abstracts and metadata such as title, authors, and publication year.

\subsection{Dataset Statistics}
Our dataset construction process results in 6{,}279 query talks linked to 43,429 cited papers in total. The query talks average 11.1 minutes in duration, and their transcripts contain on average 1{,}478 words. Each query talk is associated with an average of 26.4 cited papers, providing a rich set of references per talk.

The data split follows the original NUTSHELL dataset~\citeplanguageresource{nutshell}, which partitions talks chronologically by conference year. Talks in earlier years form the training set (2017-2021), while those from later years are used for development and testing (2022). By preserving this ordering in \textsc{Talk2Ref}, the test set consists of talks from more recent years than the training set, providing a realistic, temporally consistent evaluation scenario. Consequently, there are 3971 query talks in training, 882 query talks in development, and 1426 talks in the test split. Detailed dataset statistics are also provided in \cref{tab:dataset_stats}. 

In general, the number of references increases after 2017, mirroring real-world publishing trends in this field. Further statistics about coverage over years can be found in \cref{fig:temporal} in \cref{sec:appendix}. The ten most cited papers can be fond in  \cref{fig:top10}.


%% file: tables/dataset_stats.tex
\begin{table*}[t]
\centering
\adjustbox{max width=\textwidth}{%
\begin{tabular}{lccccccccc}
\toprule
\multirow{2}{*}{\textbf{Split}} & \multirow{2}{*}{\microphone \textbf{Confs.}} & \multirow{2}{*}{\microphone 
\textbf{Years}} & \multirow{2}{*}{\microphone \textbf{Talks}} & \multirow{2}{*}{\microphone}  \textbf{Avg. len.} &  \multirow{2}{*}{\microphone} \textbf{Avg. words/}&  \multirow{2}{*}{\docicon} \textbf{Avg.}  & \multirow{2}{*}{\docicon} \textbf{Total} & \multirow{2}{*}{\docicon} \textbf{Citation} & \multirow{2}{*}{\docicon} \textbf{Avg. words/} \\
&&&&  \textbf{(min)} & \textbf{\phantom{00}transcript} &\textbf{\phantom{0000} papers} & \textbf{\phantom{0000}papers} & \phantom{000}\textbf{years} & \textbf{abstract} \\
\midrule
\vpad Train & \makecell[c]{ACL, NAACL\\EMNLP}  & \makecell[c]{2017--\\2021} & 3971 & 12.1 & 1615.3 & 26.75 & 31{,}064 & 1948--2021 & 142.4\\
\vpad Dev   & ACL               & 2022       & 882  & 9.9  & 1326.9 & 26.05 & 11{,}805  & 1967--2022 & 147.8\\
\vpad Test  & \makecell[c]{EMNLP,\\ NAACL}      & 2022       & 1426 & 9.1  & 1186.1 & 25.66 & 16{,}935  & 1953--2022 & 149\\
\midrule
\vpad Total & \makecell[c]{ACL, NAACL\\EMNLP}  & \makecell[c]{2017--\\2022} & 6279 & 11.1 & 1477.6 & 26.4  & \makecell[c]{43{,}429} & 1948--2022 & 144.8\\
\bottomrule
\end{tabular}
}
\caption{
Dataset statistics for our proposed dataset \textbf{Talk2Ref}, that includes  \microtext{} query talks and \doctext{} cited papers. We show that citation years for years with at least 10 references, words are split at whitespace.
}
\label{tab:dataset_stats}
\end{table*}

%% file: 4_experiments.tex
\subsection{Experimental Setting}\label{sec:experimental}
\subsubsection{Input Representations}\label{subsec:input_repr}
\paragraph{\microtext Query talk representations.} 
On the query talk side, we do not use raw audio directly in the tested models, but rely on long-form transcripts of the talks. Although the transcripts can be noisy and lengthy, they enable us to leverage the rich textual content of the talk, along with its title and publication year, while keeping the input modality consistent with the cited-paper side.

Transcribing first also offers practical benefits: automatic speech recognition systems are mature and widely available, and their intermediate outputs are easier to inspect than learned audio embeddings. While emerging models such as SONAR \citep{duquenne2023sonarsentencelevelmultimodallanguageagnostic} could, in principle, support direct speech–text retrieval using similar chunking and pooling strategies, we leave this exploration to future work.

However, working with transcripts introduces another challenge: common pretrained sentence encoders like BERT \citep{bert}, SentenceBERT \citep{SentenceBERT}, or Specter \citep{specter} accept only a limited number of tokens (512), much shorter than the average transcript length of 1,478 words. To obtain fixed-size embeddings for these long transcripts, we compare several strategies:
\begin{enumerate}
    \item \emph{Truncation}: keeping only the first 512 tokens;
    \item \emph{Chunking with mean pooling}: splitting the transcript into 512-token segments and averaging their embeddings;
    \item \emph{Chunking with max pooling}: splitting the transcript into 512-token segments and taking the element-wise maximum of the segment embeddings;
    \item \emph{Chunking with learned weighted mean}: splitting the transcript into 512-token segments. A small feed-forward (linear) layer produces a scalar for each segment. The scalars are passed through a softmax layer to produce weights. These weights determine how much each segment contributes to the weighted mean. 
\end{enumerate}

We experiment with different input configurations, including using transcripts alone or appended to the query talk title and publication year.

\paragraph{\doctext Cited paper representations.} On the output side, we rely on the title, abstract, and publication year of the cited papers. We experiment with each feature individually as well as their combinations. No truncation or aggregation is needed, as these inputs fall well within the encoder context length.

\subsubsection{Training and Retrieval}\label{subsec:training}
We evaluate several models (detailed in \cref{subsubsec:models}) in both zero-shot and trained settings on our dataset. Our training strategy is detailed below.

\paragraph{Training.}  
We employ a contrastive learning setup inspired by dense passage retrieval \citep[DPR]{DPR}. We do not initialize from pretrained DPR weights due to the mismatch in task structure: DPR models are trained for open-domain question answering. Instead we use pretrained sentence embedding models as detailed below. 
Each query talk $T$ and cited paper $R_i$ is mapped to a vector representation by separate encoders $f_T$ and $f_R$, producing embeddings
$f_T(T)$ and $f_R(R_i)$, respectively. The similarity between a talk and a candidate paper is computed as the dot product $s_i = f_T(T) \cdot f_R(R_i)$. 

The model is optimized to assign higher similarity scores to correct talk–reference pairs and lower similarity scores to incorrect talk–reference pairs.

Unlike \citet{DPR}, where each query has a single positive, talks often have multiple relevant papers. We therefore replace the original softmax-based objective 
with a sigmoid-based binary classification loss, allowing multiple correct references per talk:

\begin{equation*}
    \mathcal{L} = -\sum_{i=1}^N \left[ y_i \log \sigma(s_i) + (1 - y_i) \log (1 - \sigma(s_i)) \right],
\end{equation*}

where $y_i \in \{0,1\}$ indicates whether $R_i$ is a true citation, and 
$\sigma(\cdot)$ denotes the sigmoid function. 

For efficient training, other talk–reference pairs within the same batch are used as negative examples. Specifically, for a query talk at index $i$ in the batch, all cited papers from different talks $j \neq i$ are treated as negatives, unless a cited paper is shared across both query talks.

If $f_T$ and $f_R$ have different embedding dimensions, a linear projection layer is added to align them.

\input{tables/input_repr}

\paragraph{Domain adaptation stage.}  
We also experiment with a task-specific domain adaptation stage of the pretrained encoders before the main training stage. This domain adaptation stage follows a contrastive objective similar to \citet{DPR}, however, each query talk $T$ is paired with its own abstract instead of the cited papers' abstracts. The resulting one-to-one alignment enables the use of a standard softmax-based loss, encouraging the model to learn meaningful representations before finetuning on the main citation recommendation task.

The domain adaptation stage serves two purposes. First, it adapts the encoders to the domain of scientific talks, improving their ability to process noisy, long-form transcripts. Second, it provides a simpler learning objective, as mapping a talk to its own query paper’s abstract is easier than retrieving cited papers, whose abstracts may only partially overlap with the talk.

\paragraph{Retrieval and inference.} All candidate paper embeddings are precomputed and stored in a FAISS index \citep{Faiss} for efficient retrieval. To ensure temporal consistency, we restrict retrieved papers to those published prior to the query talk paper, preventing the model from selecting “future” papers. At inference, the query talk transcript is embedded and compared against the candidate paper embeddings to identify the top-$k$ most relevant references.

\paragraph{Baselines.}  
Two baselines are considered, addressing different aspects of the task:
(i) Task-level baseline: predicting the top-$k$ most frequently cited papers across all talks, providing a simple frequency-based reference for the overall difficulty of the citation prediction task;
and
(ii) Model-level baseline: using only the first $x$ tokens of each transcript without any aggregation, serving as a lower bound for our transcript-encoding strategies.
\subsubsection{Models}\label{subsubsec:models}

To assess the effect of different pretrained encoders on the task, we evaluate four models:
\begin{itemize}
    \item  \textbf{BERT} \citep{bert}: a general-purpose language model not trained for retrieval tasks. A sentence representation is obtained by averaging over the embeddings of each token.
    
    \item  \textbf{Longformer} \citep{longformer}: a general-purpose language model capable of processing sequences of up to 4,096 tokens, thereby reducing the need for aggressive truncation.
    
    \item  \textbf{SPECTER2} \citep{specter}: a model pretrained on scientific citation data and designed to encode scientific papers, building on SciBERT \citep{SciBert}, a variant of BERT.
    
    \item  \textbf{Sentence-BERT} \citep[SBERT]{SentenceBERT}: a model optimized for producing fixed-size sentence embeddings using a siamese network architecture based on BERT.
\end{itemize}
All models have a maximum sequence length of 512 tokens, except Longformer, which can handle up to 4,096 tokens.
For BERT and Longformer, sentence embeddings are obtained by averaging token embeddings, following the approach recommended by \citet{SciBert}. In the same way, SBERT~\citep{SentenceBERT} uses mean pooling over tokens~\citep{SentenceBERT}, and SPECTER2~\citep{specter} uses the \texttt{[CLS]} token as its sentence representation, as done in the original papers.

Further details about the models, including number of parameters, are provided in \cref{tab:model_sizes} in \cref{app:encoders}.

\paragraph{Training details.}  
We train for up to 72 hours on a single GPU H100, with early stopping with patience of four epochs on validation performance. We use a batch size of $24$ for the models with sequence length of $512$, and a smaller batch size of $3$ for Longformer. 
Detailed hyperparameters are given in  \cref{tab:training_config} in \cref{sec:appendix}.

\subsubsection{Evaluation}
Following prior work on citation recommendation (\citealplanguageresource{RS_content_important}; \citealp{RS_Svm, RS_hierachical}), we evaluate retrieval quality using Precision, Precision at cutoff $k$ (P@$k$), Recall, Recall at cutoff $k$ (R@$k$), and Mean Average Precision (MAP) at cutoff  $k$ (MAP@$k$). These metrics capture complementary aspects of retrieval quality. P@$k$ and MAP@$k$ focus on the quality of the top-ranked recommendations, and R@$k$ measures the coverage of all relevant references.

We retrieve the top-$k$ most similar papers for each query talk and compute the metrics with respect to the gold set of cited papers.

We report P@$k$ and MAP@$k$ for $k \in \{10,20\}$,  R@$k$ for $k \in \{10,20, 50\}$ following conventions in previous work \citeplanguageresource{RS_content_important} and  well aligned with the characteristics of our \textsc{Talk2Ref}, where each query talk cites approximately 26 papers on average. 

%% file: tables/input_repr.tex
\begin{table*}[t]
\centering
\resizebox{\textwidth}{!}{%
\begin{tabular}{l l cc c c c c c c c}
\toprule
\microphone \textbf{Query} \textbf{Input} & 
\docicon \textbf{Key}  \textbf{Input} & 
\makecell{\textbf{Precision}} & 
\makecell{\textbf{P@} \\ \textbf{10}} & 
\makecell{\textbf{P@} \\ \textbf{20}} & 
\makecell{\textbf{R@} \\ \textbf{20}} & 
\makecell{\textbf{R@} \\ \textbf{50}} & 
\makecell{\textbf{MAP@} \\ \textbf{10}} & 
\makecell{\textbf{MAP@} \\ \textbf{20}} \\
\midrule
N/A & 10-most cited papers
& \phantom{0}10.33 & 17.48 & 11.64 & \phantom{0}9.82 & 13.82 & \underline{12.05} & \phantom{0}7.39 \\
\midrule
Transcript & Abstract
& \phantom{0}8.89 & 12.14 & \phantom{0}9.65 & \phantom{0}7.79 & 13.18 & \phantom{0}6.63 & \phantom{0}4.49 \\
Transcript & Abstract + Title 
& \phantom{0}9.32 & 12.82 & 10.02 & \phantom{0}8.16 & 13.74 & \phantom{0}7.12 & \phantom{0}4.77 \\


Transcript & Abstract + Title + Year 
& \phantom{0}9.32 & 12.79 & 10.07 & \phantom{0}8.21 & 13.76 & \phantom{0}7.12 & \phantom{0}4.78 \\
Transcript + Title & Abstract + Title + Year 
& \phantom{0}9.55 & \textbf{13.32} & 10.39 & \phantom{0}8.58 & 14.13 & \phantom{0}7.15 & \phantom{0}4.83 \\
Transcript  + Title + Year & Abstract + Title + Year 
& \phantom{0}\textbf{9.57} & 13.29 & \textbf{10.42} & \phantom{0}\textbf{8.59} & \textbf{14.21 }& \phantom{0}\textbf{7.33} & \phantom{0}\textbf{4.94} \\
\midrule
Abstract & Abstract + Title + Year 
& \textbf{\underline{13.19}} & \textbf{\underline{19.02}} & \textbf{\underline{14.54}} & \textbf{\underline{12.05}} & \textbf{\underline{18.99}} & \textbf{11.05} & \phantom{0}\textbf{\underline{7.50}} \\
Abstract + Title & Abstract + Title + Year 
& 13.06 & 18.73 & 14.40 & 11.97 & 18.83 & 10.95 & \phantom{0}7.45 \\
Abstract + Title + Year & Abstract + Title + Year 
& 13.03 & 18.70 & 14.29 & 11.86 & 18.82 & 10.88 & \phantom{0}7.39 \\
\bottomrule
\end{tabular}%
}
\caption{Results (in $\%$) on the Talk2Ref dataset using zero-shot SBERT, truncating inputs to SBERT’s maximum sequence length. We experiment with different features for both query talks and candidate papers (keys). Using abstracts on the key side is an unrealistic setting, as abstracts are not provided with the talk; we include them to contextualize the difficulty of the task.}
\label{tab:zeroshot_results_main}
\end{table*}

%% file: 5_results.tex
\subsection{Results}\label{sec:results}
\input{tables/main_results}
In the following, we report results on \textsc{Talk2Ref} and analyze different aspects of the task, including which input features to use, the impact of using full talk transcripts versus only abstracts, and the effects of different fine-tuning strategies.

\paragraph{Selecting input features.}
We begin by exploring various input configurations in the zero-shot setting to assess which features contribute most to retrieval quality, using SBERT as the encoder. The results are reported in \cref{tab:zeroshot_results_main}. Starting with transcripts for the query and abstracts for the candidate papers (keys), we incrementally add title and year information on both sides. We find that including these additional features consistently improves performance, and adopt the final configuration for subsequent experiments. However, this zero-shot model is still surpassed by the simple but strong baseline of always predicting the most frequently cited papers in the dataset (first row). 
Results on more input combinations and higher $k$ for recall can be found in \cref{tab:input_repr_app} in \cref{app:results}.

\paragraph{Contextualizing Task Difficulty: Spoken vs. Textual Content} We evaluate an alternative setting in which the long-form, noisy transcript is replaced by the abstract of the talk’s corresponding paper (\cref{tab:zeroshot_results_main}). While this is an unrealistic scenario, since talks often do not provide abstracts, it helps contextualize the relative difficulty of retrieving references from spoken versus textual content. We find that using abstracts instead of transcripts significantly improves performance, showing that state-of-the-art zero-shot models struggle with transcripts as input. This underscores the need for our newly introduced \textsc{Talk2Ref} dataset.

\paragraph{Evaluating different SOTA models.}
Next, we explore which encoder model is most suitable for our task and can be used for subsequent training, avoiding the need to train all models for efficiency reasons. These results can be found in the top lines in \cref{tab:training_scratch}.  Unsurprisingly, a clear difference can be observed between models specifically trained to produce meaningful representations for scientific documents, such as SPECTER2 and SBERT, and general-purpose language models. Among the specialized models, SBERT outperforms SPECTER2 and is therefore chosen for the remaining experiments. We also include Longformer due to its ability to handle longer input sequences, eliminating the need for truncation or aggregation to fit the transcript. While its out-of-the-box performance is lower, we expect it to benefit from fine-tuning on our retrieval task.

Results for higher recall $k$ can be found in \cref{tab:model_comparison} in \cref{sec:appendix}, these do not change the ranking of the models.

\paragraph{Performance of finetuned models.}
We now finetune SBERT and Longformer on \textsc{Talk2Ref}.
Since talk transcripts exceed SBERT’s input limits, we apply chunking and aggregation strategies on the query side as detailed in \cref{subsec:input_repr}: truncation, mean pooling, and a learnable weighted mean. On the key side (paper abstracts, titles, and years), no aggregation is required due to the shorter lengths.

For Longformer, by contrast, the full transcript fits within the model’s context window, so no chunking or aggregation is needed on the query side. However, on the key side, we continue to use SBERT as the encoder, since it provides substantially stronger representations for abstracts than Longformer as discussed above.

Results for these trained models are shown in the middle section of \Cref{tab:training_scratch}. 
Unsurprisingly, training on \textsc{Talk2Ref} substantially improves performance for both SBERT and Longformer, with SBERT now outperforming the top-$k$ most frequent baseline. Among the different aggregation strategies, the learned weighted mean performs best for SBERT, surpassing truncation and simple mean pooling. 

For Longformer, finetuning yields a substantial improvement, though the model still performs slightly worse than zero-shot SBERT, likely because SBERT has been explicitly trained for representation learning, whereas Longformer has not. We additionally experimented with splitting the input and applying aggregation mechanisms for Longformer as well; however, these configurations did not lead to further gains. 
The corresponding results are reported in \cref{tab:finetuning_results_app} in \cref{app:results}.

\paragraph{Effect of the domain adaptation stage.}
Finally, we add a task-specific domain adaptation stage to our models before finetuning, as described in \cref{subsec:training}. This stage serves two purposes: it provides a simpler learning task and helps the model adapt to the scientific domain. We include two configurations in our evaluation: SBERT (learned mean), the best-performing model after fine-tuning, and Longformer/SBERT, which shows significant gains from fine-tuning. 

The bottom rows of \Cref{tab:training_scratch} show that domain adaptation further improves performance, with only small gains for Longformer/SBERT but significant improvements for SBERT (learned mean). In comparison to SBERT's finetuned model with learned mean (but without domain adaptation stage), it gains improvements across all metrics. In fact, it achieves the highest scores across all metrics except \textsc{MAP@10}, where the frequency-based baseline remains superior. These results indicate that the adaptation stage enhances the model’s ability to capture semantic relevance beyond surface-level similarity.

\subsection{Discussion}
\paragraph{Challenges of \textsc{Talk2Ref}.}
The \textsc{Talk2Ref} dataset presents several challenges that make reference prediction from talks a non-trivial task. First, reference prediction from talks is significantly harder than from papers or their abstracts, as shown in \cref{tab:zeroshot_results_main}. Second, transcripts are long and often exceed the input window of standard encoders, making the production of fixed-size embeddings challenging. Aggregation strategies, such as mean or weighted pooling over chunks, help capture information from longer inputs, as illustrated in \cref{tab:training_scratch}, though they provide only an approximate representation.

Our experiments demonstrate that finetuning on \textsc{Talk2Ref} substantially improves retrieval performance, showing that the dataset provides rich and meaningful training signals. The results also highlight the benefits of both aggregation strategies and the domain adaptation stage, confirming that \textsc{Talk2Ref} supports effective learning for the task of reference prediction from scientific talks.

\paragraph{Comparison to prior work.}  
Prior work on citation recommendation, such as \citetlanguageresource{RS_content_important}, achieves competitive results using clean, structured text (titles and abstracts) on both the query and key sides. For example, their NNSelect model reaches \( P@10 = 28.7 \), \( P@20 = 23.0 \), and \( R@20 = 36.3 \) on the DBLP dataset \citeplanguageresource{dblp_acm}.   In comparison, our best finetuned model on \textit{\textsc{Talk2Ref}} achieves \( P@10 = 19.1 \), \( P@20 = 15.3 \), and \( R@20 = 12.7 \). While these numbers are lower, they remain in a comparable range, highlighting that our task, predicting references from long, noisy spoken-language transcripts, is substantially more challenging. Nonetheless, the results demonstrate that our dual-encoder models can still produce reasonable and meaningful predictions in this difficult setting.

%% file: tables/main_results.tex
\begin{table*}[t]
\centering
\resizebox{\textwidth}{!}{%
\begin{tabular}{l l l c l c c c c c c c}
\toprule
\textbf{Strategy} & 
\makecell{\microphone \docicon\\ \textbf{Encoder}} & 
\makecell{\microphone \docicon \\\textbf{Max} \\ \textbf{Seq. Len}} & 
\makecell[c]{\microphone \\\textbf{Aggregation}} &
\makecell{\textbf{Prec.}} & 
\makecell{\textbf{P@} \\ \textbf{10}} & 
\makecell{\textbf{P@} \\ \textbf{20}} & 
\makecell{\textbf{R@} \\ \textbf{20}} & 
\makecell{\textbf{R@} \\ \textbf{50}} & 
\makecell{\textbf{MAP@} \\ \textbf{10}} & 
\makecell{\textbf{MAP@} \\ \textbf{20}} \\
\midrule

Top-k most freq. & -- & -- & -- 
& 10.33 & 17.48 & 11.63 & 9.82 & 13.81 & \textbf{12.04} & \phantom{0}7.39 \\
\midrule
\multirow{5}{*}{Zero-Shot} & BERT & 512 & Truncated 
& \phantom{0}0.56 & \phantom{0}0.67 & \phantom{0}0.57 & \phantom{0}0.49 & \phantom{0}0.82 & \phantom{0}0.24 & \phantom{0}0.17 \\

 & Longformer & 4096  & -- & \phantom{0}0.20 & \phantom{0}0.23 & \phantom{0}0.21 & \phantom{0}0.18 &\phantom{0}0.34 & \phantom{0}0.08 &\phantom{0}0.05\\

 & Specter2 & 512 & Truncated 
& \phantom{0}8.31 & 11.78 & \phantom{0}9.10 & \phantom{0}7.37 & 12.43 & \phantom{0}6.15 & \phantom{0}4.09 \\

 & SBERT & 512 & Truncated 
& \phantom{0}9.57 & 13.29 & 10.42 & \phantom{0}8.59 & 14.21 & \phantom{0}7.33 & \phantom{0}4.94 \\
\cmidrule{2-11}
 & SBERT & 512 & Mean 
& 10.60 & 14.83 & 11.71 & \phantom{0}9.54 & 15.76 & \phantom{0}8.21 & \phantom{0}5.56 \\
\midrule

\multirow{5}{*}{Finetuned} & SBERT & 512 & Truncated 
& 12.94 & 17.23 & 13.85 & 11.49 & 19.59 & 9.76 & \phantom{0}6.91 \\

 & SBERT & 512 & Mean 
& 13.34 & 17.75 & 14.28 & 11.92 & 20.35 & 10.13 & \phantom{0}7.15 \\

 & SBERT & 512 & Learned mean 
& 13.95 & 18.81 & 15.12 & 12.59 & 21.00 & 10.86 & \phantom{0}7.69 \\
\cmidrule{2-11}
& \makecell[l]{\microtext Longformer/\\ \doctext SBERT} & 4096 & -- 
& \phantom{0}8.47 &  10.75 & \phantom{0}9.15 & \phantom{0}7.32 & 14.06 &  \phantom{0}5.12 & \phantom{0}3.57 \\
\midrule
\multirow{3}{*}{\makecell[l]{Dom. Adapt. \\+ Finetuned}}  & SBERT & 512 & Learned mean 
& \textbf{14.18} & \textbf{19.14} & \textbf{15.32} & \textbf{12.71} & \textbf{21.60} & 10.98 & \phantom{0}\textbf{7.72} \\

& \makecell[l]{\microtext Longformer/\\ \doctext SBERT} & 4096 & -- 
& \phantom{0}8.90 & 11.25  & \phantom{0}9.57   & \phantom{0}7.28 & 14.16 & \phantom{0}5.43 & \phantom{0}3.78 \\
\bottomrule

\end{tabular}%
}
\caption{Results (in $\%$) on the Talk2Ref dataset. Query inputs consist of transcript, title, and year, while cited papers (keys) are represented by title, year, and abstract. We report different aggregation mechanisms for handling long query transcripts apart from Longformer, where the transcript fits completely, and show results for finetuning the best zero-shot model on Talk2Ref.}
\label{tab:training_scratch}
\end{table*}

%% file: appendix.tex
\appendix\label{sec:appendix}

\section{Dataset Statistics}
\begin{figure*}[ht]
    \centering
    \includegraphics[width=0.8\linewidth]{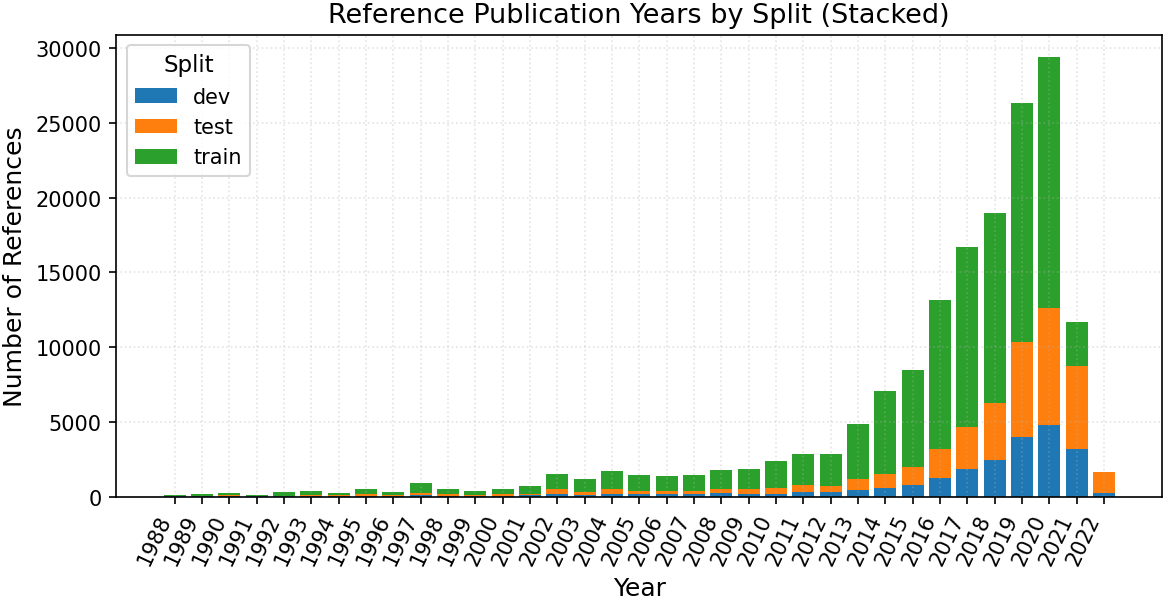}
    \caption{Temporal distribution of cited works and abstracts in the dataset. The majority of references are concentrated between 2015 and 2022, with a marked increase from 2018 onward, reflecting the surge of research in natural language processing. This distribution ensures alignment with current research trends but underrepresents older foundational work.}
    \label{fig:temporal}
\end{figure*}

\begin{figure*}[ht]
    \centering
    \includegraphics[width=\textwidth, keepaspectratio]{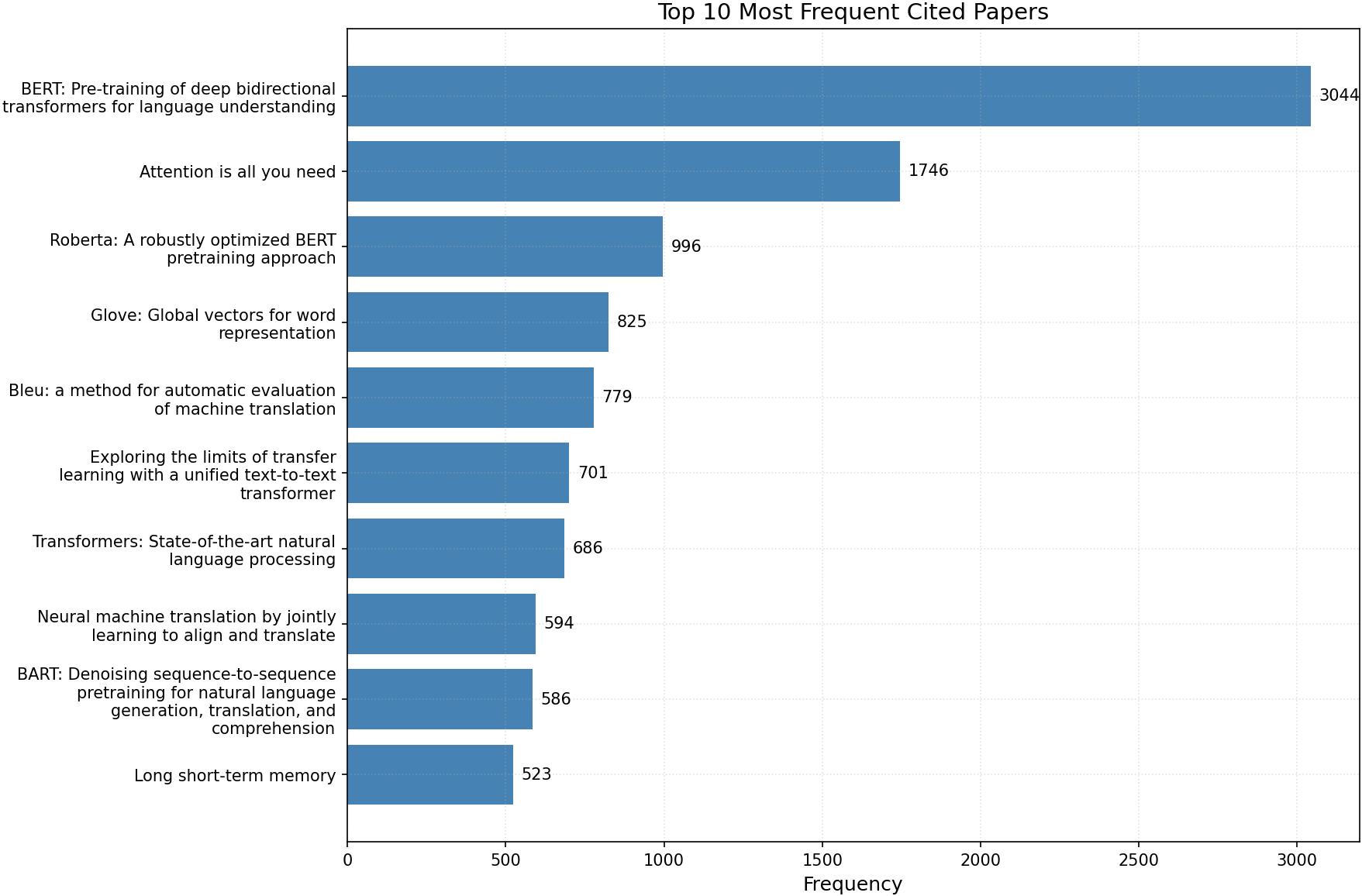}
    \caption{Top 10 most frequently cited papers in the dataset.}
    \label{fig:top10}
\end{figure*}

The temporal coverage of references is concentrated in recent years: the majority of all cited works were published between 2015 and 2022. A clear growth trend is visible, with particularly sharp increases in references from 2018 onwards, mirroring the surge of research activity in natural language processing, with newer publications dominating the dataset as cited papers. This temporal skew ensures alignment with current research but naturally reduces representation of older foundational works. The distribution over years is shown  in \cref{fig:temporal}. The ten most cited papers in Talk2Ref are shown in \cref{fig:top10}.

\section{Encoder Models}\label{app:encoders}
Detailed information to the encoder models used in this work can be found in \cref{tab:model_sizes}. Hyperparmaters used to train these models are given in \cref{tab:training_config}.

\begin{table}[ht]
\centering
\resizebox{\linewidth}{!}{%
\begin{tabular}{llp{1cm}cc}
\toprule
\textbf{Model} & \textbf{HF-ID} & \textbf{Ref.} & \textbf{\# Params} & \textbf{\makecell{Transformer\\Version}} \\
\midrule
SBERT & \makecell[l]{\texttt{sentence-transformers/}\\\texttt{all-MiniLM-L6-v2}} & \citet{SentenceBERT} & 22.7M & 4.51.3 \\
Longformer & \makecell[l]{\texttt{allenai/}\\\texttt{longformer-base-4096}} & \citet{longformer} & 148.8M & 4.51.3 \\
Specter & \makecell[l]{\texttt{allenai/}\\\texttt{specter2\_base}} & \citet{specter} & 110M & 4.51.3 \\
BERT & \texttt{bert-base-uncased} & \citet{bert} & 110M & 4.51.3 \\
\bottomrule
\end{tabular}%
}
\caption{%
Encoder models used in our experiments. All models use Transformers v4.51.3. 
For zero-shot experiments, the same encoder is used for both query and key encoding without any aggregation or projection layers. 
During training, SBERT~\citep{SentenceBERT} or Longformer~\citep{longformer} serve as query encoders, while the key encoder remains fixed to SBERT. 
A linear projection layer adds approximately 0.3M parameters when applied, and a learned weighted mean aggregation layer adds about 0.04M (SBERT) or 0.15M (Longformer) parameters on the query side.%
}
\label{tab:model_sizes}
\end{table}


\begin{table}[ht]
\centering
\resizebox{\linewidth}{!}{%
\begin{tabular}{l c c}
\toprule
\textbf{Parameter} & \\
\midrule
\microtext Query Encoder & SBERT & Longformer \\
\doctext Key Encoder   & SBERT & SBERT \\
Batch size         & 24 & 3 \\
Max.\ Epochs       & 100 & 100 \\
Freeze layers (Key side) & 2 & 4 \\
Freeze layers (Query side) & 2 & 8 \\
Gradient accumulation & 3 & 3 \\
Learning rate (base) & 6e-6 & 6e-6 \\
Learning rate (head) & 2e-4 & 2e-4 \\
Weight decay       & 0.01 & 0.01 \\
Dropout rate       & 0.05 & 0.05 \\
Adam epsilon       & 1e-8 & 1e-8 \\
Avg. inference time (s / example) & 0.0083 & 0.0286 \\
Early Stopping in Epochs & 4 & 4 \\
\bottomrule
\end{tabular}
}
\caption{Training configurations for finetuning SBERT \citep{SentenceBERT} and Longformer\citep{longformer} on the Talk2Ref dataset.}
\label{tab:training_config}
\end{table}

\section{Results and Discussion}\label{app:results}

This section provides an extended overview of the experimental results and additional analyses that complement the findings presented in the main paper.

\subsection{Input representations}

\input{tables/input_repr_app}

To determine the most effective input configuration, we evaluate all combinations of query and key inputs using SBERT~\citep{SentenceBERT} as encoder (\cref{tab:input_repr_app}). On the key and query side, the combination of \textit{abstract, title, and year} yields the best results across nearly all metrics and is therefore fixed for the remaining experiments.

However, the non-realistic scenario of using text abstracts instead of transcripts, remain markedly stronger: Abstract+Title+Year outperforms Transcript+Title+Year by \(\sim\)5.43 P@10 points (18.72 vs.\ 13.29) and \(\sim\)5.44 R@100 points (25.43 vs.\ 19.99). This gap is consistent with transcripts being long, noisy, and truncated to 512 tokens, whereas abstracts are concise, well-structured summaries. Since our goal is citation recommendation from the talk, we treat abstract-based results as a strong upper-bound baseline and focus modeling efforts on transcript-based inputs because talks rarely provide abstracts or summaries.

\subsection{Model Selection}

\input{tables/diff_models_app}

Having fixed the input representation, we next compare encoder architectures to identify the most suitable model for subsequent training (\cref{tab:model_comparison}). 
Across all configurations, SBERT~\citep{SentenceBERT} consistently outperforms other encoders by a clear margin, achieving the highest precision and recall values for both abstract and transcript inputs (e.g., P@10 = 18.70 vs.\ 15.42 for SPECTER2\citep{specter} and 13.29 for transcript-based SBERT). SPECTER2 performs second-best, followed by Longformer \cite{longformer} and BERT \citep{bert}, which struggle under the truncated input. Based on these results, we fix SBERT with the full input configuration (transcript + title + year on the query side; title + year + abstract on the key side) for all subsequent training experiments to ensure efficiency and consistency.

\newpage
\input{tables/finetuning_results_app}
\subsection{Finetuned Models}\label{sec:app_train_model}
After identifying SBERT~\citep{SentenceBERT} as the most effective encoder and fixing the input representation (\textit{Transcript + Title + Year} on the query side, \textit{Title + Year + Abstract} on the key side), we trained dual-encoder models under different regimes: on training task, on the domain adaptation stage, and with subsequent training on domain adapted checkpoints.  We also include Longformer \citep{longformer} due to its capacity for longer sequences; although its out-of-the-box performance is lower, we expect it to benefit from fine-tuning on our retrieval task. 
Results are summarized in \cref{tab:finetuning_results_app}.

When finetuning, SBERT-based models consistently outperform all baselines. Among aggregation strategies, the \textit{learned weighted mean} yields the strongest results (\(P@20 = 15.12\), \(R@200 = 38.21\)), clearly surpassing both simple mean pooling and truncation. This indicates that weighting informative segments within long transcripts allows the model to better capture semantic relations between talks and their cited papers. Compared to the most-cited baseline (\(R@200 = 21.82\)), recall improves by more than 70\%, while precision gains remain moderate—suggesting that the model retrieves semantically related papers even when not all are explicitly cited.

Finetuning on the continual pretrained SBERT \citep{SentenceBERT} checkpoint further improves results, with the learned weighted mean configuration achieving the overall best performance (\(P@20 = 15.31\), \(R@200 = 39.04\)). This confirms that the domain adaptation stage facilitates better representation learning for long-form transcripts and that finetuning adapts these representations effectively to the citation retrieval task.

Models using the Longformer~\citep{longformer} encoder perform notably worse, limited by GPU memory constraints and small batch sizes (3 vs.\ 24 for SBERT). Mean aggregation slightly outperforms the learned weighted mean, implying that the model struggles to learn attention weights effectively across extended contexts. Although recall at larger cutoffs (e.g., \(R@200 \approx 29.82\)) exceeds the frequency baseline, Precision and MAP values remain significantly lower. These results suggest that, despite its architectural suitability for long sequences, Longformer \citep{longformer} requires larger datasets and more compute to fully leverage its capacity.

%% file: tables/input_repr_app.tex
\begin{table*}[t]
\centering
\resizebox{\textwidth}{!}{%
\begin{tabular}{l l p{2.6cm} p{3.0cm} c c c c c c c c c}
\toprule
\makecell{\microphone \textbf{Query}  \textbf{Input}} & 
\makecell{\docicon \textbf{Key}  \textbf{Input}} & 
\makecell{\textbf{Prec.}} & 
\makecell{\textbf{P@} \\ \textbf{10}} & 
\makecell{\textbf{P@} \\ \textbf{20}} & 
\makecell{\textbf{R@} \\ \textbf{20}} & 
\makecell{\textbf{R@} \\ \textbf{50}} & 
\makecell{\textbf{R@} \\ \textbf{100}} & 
\makecell{\textbf{R@} \\ \textbf{200}} & 
\makecell{\textbf{MAP@} \\ \textbf{10}} & 
\makecell{\textbf{MAP@} \\ \textbf{20}} \\
\midrule

N/A &  most cited papers 
& 10.33 & 17.48 & 11.64 &  9.82 & 13.82 & 17.25 & 21.83 & \textbf{\underline{12.05}} &  7.39 \\
\midrule
  \makecell{Transcript} & \makecell{Abstract}
&  \phantom{0}8.89 & 12.14 &  \phantom{0}9.65 &  \phantom{0}7.79 & 13.18 & 18.87 & 25.49 &  \phantom{0}6.63 &  \phantom{0}4.49 \\
\makecell{Transcript} & \makecell{Title}
&  \phantom{0}8.32 & 11.40 &  \phantom{0}8.94 &  \phantom{0}7.33 & 12.30 & 17.41 & 23.70 &  \phantom{0}6.10 &  \phantom{0}4.10 \\
\makecell{Transcript} & \makecell{Title  + Abstract}
&  \phantom{0}9.32 & 12.82 & 10.02 &  \phantom{0}8.16 & 13.74 & 19.56 & 26.08 &  \phantom{0}7.12 &  \phantom{0}4.77 \\
\makecell{Transcript} & \makecell{Abstract  + Year}
&  \phantom{0}8.87 & 12.12 &  \phantom{0}9.59 &  \phantom{0}7.74 & 13.12 & 18.77 & 25.36 &  \phantom{0}6.68 &  \phantom{0}4.50 \\
 \makecell{Transcript} & \makecell{Title  + Year}
&  \phantom{0}7.61 & 10.60 &  \phantom{0}8.22 &  \phantom{0}6.78 & 11.22 & 15.95 & 21.72 &  \phantom{0}5.46 &  \phantom{0}3.60 \\
 \makecell{Transcript} & \makecell{Title + Year  + Abstract}
&  \phantom{0}9.32 & 12.79 & 10.07 &  \phantom{0}8.21 & 13.76 & 19.46 & 26.10 &  \phantom{0}7.12 &  \phantom{0}4.78 \\
 \makecell{Transcript  + Title} & \makecell{Title + Year  + Abstract}
&  \phantom{0}9.55 & \textbf{13.32} & 10.39 &  \phantom{0}8.58 & 14.13 & 19.87 & 26.65 &  \phantom{0}7.15 &  \phantom{0}4.83 \\
 \makecell{Transcript  + Year} & \makecell{Title + Year  + Abstract}
&  \phantom{0}9.45 & 13.07 & 10.27 &  \phantom{0}8.42 & 14.04 & 19.81 & 26.60 &  \phantom{0}7.25 &  \phantom{0}4.86 \\
 \makecell{Transcript  + Title  + Year} & \makecell{Title + Year  + Abstract}
&  \textbf{\phantom{0}9.57} & 13.29 & \textbf{10.42} &  \textbf{\phantom{0}8.59} & \textbf{14.21} & \textbf{19.99} & \textbf{26.72} &  \textbf{\phantom{0}7.33} &  \textbf{\phantom{0}4.94} \\
\midrule
\makecell{Title} & \makecell{Title + Year  + Abstract}
&  \phantom{0}9.95 & 13.56 & 10.69 &  \phantom{0}9.09 & 14.60 & 19.92 & 26.56 &  \phantom{0}7.52 &  \phantom{0}5.21 \\
 \makecell{Title  + Year} & \makecell{Title + Year  + Abstract}
&  \phantom{0}9.68 & 12.92 &  \phantom{0}8.81 & 10.28 & 14.22 & 19.50 & 25.99 &  \phantom{0}7.19 &  \phantom{0}5.02 \\
 \makecell{Abstract} & \makecell{Title + Year  + Abstract}
& \textbf{\underline{13.19}} & \textbf{\underline{19.02}} & \textbf{\underline{14.54}} & \textbf{\underline{12.05}} & \textbf{\underline{18.99}} & \textbf{\underline{25.75}} & \textbf{\underline{33.06}} & \textbf{11.05} & \textbf{\underline{\phantom{0}7.50}} \\
\makecell{Abstract  + Title} & \makecell{Title + Year  + Abstract}
& 13.06 & 18.73 & 14.40 & 11.96 & 18.83 & 25.61 & 32.80 & 10.95 &  \phantom{0}7.45 \\
\makecell{Abstract  + Title  + Year} & \makecell{Title + Year  + Abstract}
& 13.03 & 18.70 & 14.30 & 11.86 & 18.82 & 25.43 & 32.67 & 10.88 &  \phantom{0}7.39 \\
\bottomrule
\end{tabular}%
}
\caption{Results on the Talk2Ref dataset using zero-shot SBERT, truncating inputs to SBERT’s maximum sequence length. We experiment with different features for both query talks and candidate papers (keys). Using abstracts on the key side is an unrealistic setting, as abstracts are not provided with the talk; we include them to contextualize the difficulty of the task.}
\label{tab:input_repr_app}
\end{table*}

%% file: tables/diff_models_app.tex
\begin{table*}[t]
\centering
\resizebox{\textwidth}{!}{%
\begin{tabular}{l l c l c c c c c c c c c}
\toprule
\textbf{Strategy} & 
\makecell{\microphone \docicon\\ \textbf{Encoder}} & 
\makecell{\microphone \docicon\\ \textbf{Max} \\ \textbf{Seq. Len}} & 
\makecell[c]{\microphone\\ \textbf{Aggregation}} &
\makecell{\textbf{Prec.}} & 
\makecell{\textbf{P@}\\ \textbf{10}} & 
\makecell{\textbf{P@}\\ \textbf{20}} & 
\makecell{\textbf{R@}\\ \textbf{20}} & 
\makecell{\textbf{R@}\\ \textbf{50}} & 
\makecell{\textbf{R@}\\ \textbf{100}} & 
\makecell{\textbf{R@}\\ \textbf{200}} & 
\makecell{\textbf{MAP@}\\ \textbf{10}} & 
\makecell{\textbf{MAP@}\\ \textbf{20}} \\
\midrule
Top-k most freq. & -- & -- & -- 
& \underline{10.33} & \underline{17.48} & \underline{11.64} & \underline{\phantom{0}9.82} & 13.82 & 17.25 & 21.83 & \underline{12.05} & \underline{\phantom{0}7.39} \\
\midrule
\multirow{4}{*}{Zero-Shot} 
 & SBERT & 512 & Truncated 
& \textbf{\phantom{0}9.57} & \textbf{13.29} & \textbf{10.42} & \textbf{\phantom{0}8.59} & \textbf{\underline{14.21}} & \textbf{\underline{19.99}} & \textbf{\underline{26.72}} & \textbf{\phantom{0}7.33} & \textbf{\phantom{0}4.94} \\

 & Specter2 & 512 & Truncated 
& \phantom{0}8.31 & 11.78 & \phantom{0}9.10 & \phantom{0}7.37 & 12.43 & 17.39 & 23.81 & \phantom{0}6.15 & \phantom{0}4.09 \\

 & Longformer & 4096 & -- 
& \phantom{0}0.17 & \phantom{0}0.21 & \phantom{0}0.20 & \phantom{0}0.16 & \phantom{0}0.31 & \phantom{0}0.49 & \phantom{0}0.97 & \phantom{0}0.10 & \phantom{0}0.06 \\

 & BERT & 512 & Truncated 
& \phantom{0}0.56 & \phantom{0}0.67 & \phantom{0}0.57 & \phantom{0}0.49 & \phantom{0}0.82 &  \phantom{0}1.43 &  2.32 & \phantom{0}0.24 & \phantom{0}0.17 \\
\bottomrule
\end{tabular}%
}
\caption{Results on the Talk2Ref dataset for zero-shot retrieval. Query inputs consist of transcript, title, and year, while cited papers (keys) are represented by title, year, and abstract.}
\label{tab:model_comparison}
\end{table*}

%% file: tables/finetuning_results_app.tex
\begin{table*}[ht]
\resizebox{\textwidth}{!}{%
\begin{tabular}{l l l c l c c c c c c c c c}
\toprule
\textbf{Strategy} & 
\makecell{\microphone  \\\textbf{Query} \\ \textbf{Enc.}} & 
\makecell{\docicon \\\textbf{Key} \\ \textbf{Enc.}} & 
\makecell{\microphone \docicon \\\textbf{Max} \\ \textbf{Seq.}} & 
\makecell{\microphone  \\\textbf{Aggregation}} &
\makecell{\textbf{Prec.}} & 
\makecell{\textbf{P@} \\ \textbf{10}} & 
\makecell{\textbf{P@} \\ \textbf{20}} & 
\makecell{\textbf{R@} \\ \textbf{20}} & 
\makecell{\textbf{R@} \\ \textbf{50}} & 
\makecell{\textbf{R@} \\ \textbf{100}} & 
\makecell{\textbf{R@} \\ \textbf{200}} & 
\makecell{\textbf{MAP@} \\ \textbf{10}} & 
\makecell{\textbf{MAP@} \\ \textbf{20}} \\
\midrule

Top-k most freq. & -- & -- & -- & \makecell{10-most \\ cited papers}
& 10.33 & 17.48 & 11.64 & \phantom{0}9.82 & 13.82 & 17.25 & 21.83 & \underline{12.05} & 7.39 \\
\midrule
Zero-shot & SBERT & SBERT & 512 & Truncated 
& \phantom{0}9.57 & 13.29 & 10.42 & \phantom{0}8.59 & 14.21 & 19.99 & 26.72 & \phantom{0}7.33 & 4.94 \\
Zero-shot & Longformer & Longformer & 4096 & -- 
& \phantom{0}0.20 & \phantom{0}0.23 & \phantom{0}0.21 & \phantom{0}0.18 &\phantom{0}0.34 & \phantom{0}0.59 & \phantom{0}0.109&\phantom{0}0.08 & \phantom{0}0.05  \\
\midrule
\multirow{3}{*}{Finetuned} 
 & SBERT & SBERT & 512 & Truncated 
& 12.94 & 17.23 & 13.85 & 11.49 & 19.59 & 27.34 & 35.88 & \phantom{0}9.76 & 6.91 \\
 & SBERT & SBERT & 512 & \makecell{Learned mean} 
& \textbf{13.95} & \textbf{18.81} & \textbf{15.12} & \textbf{12.59} & \textbf{20.97} & \textbf{29.18} & \textbf{38.21} & \textbf{10.86} & \textbf{7.69} \\
 & SBERT & SBERT & 512 & Mean 
& 13.34 & 17.75 & 14.28 & 11.92 & 20.35 & 28.01 & 36.65 & 10.13 & 7.15 \\
\midrule
\multirow{6}{*}{Finetuned} 
 & Longformer & SBERT & 1024 & Truncated 
& \phantom{0}8.89 & 11.19 & \phantom{0}9.53 & \phantom{0}7.67 & 14.56 & 22.06 & 31.65 & \textbf{\phantom{0}5.35} & \textbf{3.73} \\
 & Longformer & SBERT & 2048 & Truncated 
& \textbf{\phantom{0}8.95} & \textbf{11.20} & \textbf{\phantom{0}9.68} & \textbf{\phantom{0}7.74} & \textbf{14.58} & \textbf{22.21} & \textbf{31.72} & \phantom{0}5.32 & 3.72 \\
 & Longformer & SBERT & 4096 & -- 
& \phantom{0}8.47 & 10.75 & \phantom{0}9.15 & \phantom{0}7.32 & 14.06 & 21.41 & 30.68 & \phantom{0}5.12 & 3.57 \\
 & Longformer & SBERT & 512 & Mean 
& \phantom{0}8.35 & 10.37 & \phantom{0}8.90 & \phantom{0}7.16 & 13.70 & 21.04 & 29.82 & \phantom{0}4.57 & 3.27 \\
 & Longformer & SBERT & 1024 & Mean 
& \phantom{0}8.14 & 10.18 & \phantom{0}8.79 & \phantom{0}7.00 & 13.34 & 20.44 & 29.32 & \phantom{0}4.61 & 3.27 \\
 & Longformer & SBERT & 512 & \makecell{Learned mean} 
& \phantom{0}7.70 & \phantom{0}9.59 & \phantom{0}8.19 & \phantom{0}6.57 & 12.78 & 19.89 & 28.86 & \phantom{0}4.45 & 3.08 \\
\midrule
\multirow{2}{*}{Domain Adapt.} 
 & SBERT & SBERT & 512 & Truncated 
& \textbf{11.36} & \textbf{16.06} & \textbf{12.36} & \textbf{10.37} & \textbf{16.52} & \textbf{22.41} & \textbf{29.30} & \textbf{\phantom{0}9.04} & \textbf{6.18} \\
 & SBERT & SBERT & 512 & \makecell{Learned mean} 
& \phantom{0}8.79 & 12.34 & \phantom{0}9.67 & \phantom{0}7.90 & 12.76 & 17.93 & 24.03 & \phantom{0}6.65 & 4.49 \\
\midrule
Domain Adapt. + Finetuned & SBERT & SBERT & 512 & \makecell{Learned mean} 
& \textbf{\underline{14.18}} & \textbf{\underline{19.14}} & \textbf{\underline{15.32}} & \textbf{\underline{12.71}} & \textbf{\underline{21.60}} & \textbf{\underline{29.92}} & \textbf{\underline{39.04}} & 10.98 & \textbf{\underline{7.72}} \\

\bottomrule
\end{tabular}%
}
\caption{Results on the Talk2Ref dataset. Query inputs consist of transcript, title, and year, while cited papers (keys) are represented by title, year, and abstract. We report different aggregation mechanisms for handling long query transcripts and show results for finetuning the best zero-shot model on Talk2Ref.}
\label{tab:finetuning_results_app}
\end{table*}